\documentclass[conference]{IEEEtran}
\usepackage{times}

\usepackage[numbers]{natbib}
\usepackage{multicol}
\usepackage[bookmarks=true]{hyperref}

\usepackage{cite}
\usepackage{amsmath,amssymb,amsfonts}
\usepackage{algorithmic}
\usepackage{graphicx}
\usepackage{textcomp}
\usepackage{xcolor}

\newcommand{\ph}[1]{{\textbf{#1:}}}

\begin{document}
% paper title
% \title{Radio Propagation Environment Modeling and Learning for Networked Multi-Robot Exploration}
\title{PropEM-L: Radio Propagation Environment Modeling and Learning for Communication-Aware Multi-Robot Exploration}

\author{\authorblockN{Lillian Clark\authorrefmark{1},
Jeffrey A. Edlund\authorrefmark{2},
Marc Sanchez Net\authorrefmark{2},
Tiago Stegun Vaquero\authorrefmark{2}, and
Ali-akbar Agha-mohammadi\authorrefmark{2}}

\authorblockA{\authorrefmark{1}Ming Hsieh Department of Electrical and Computer Engineering,
University of Southern California\\
% Email: lilliamc@usc.edu
}
\authorblockA{\authorrefmark{2}NASA Jet Propulsion Laboratory, 
California Institute of Technology\\
Emails: lilliamc@usc.edu, \{jeffrey.a.edlund, marc.sanchez.net, tiago.stegun.vaquero, aliakbar.aghamohammadi\} @jpl.nasa.gov}
}

\maketitle

\begin{abstract}
    Multi-robot exploration of complex, unknown environments benefits from the collaboration and cooperation offered by inter-robot communication.
    Accurate radio signal strength prediction enables communication-aware exploration.
    Models which ignore the effect of the environment on signal propagation or rely on \textit{a priori} maps suffer in unknown, communication-restricted (e.g. subterranean) environments.
    In this work, we present Propagation Environment Modeling and Learning (PropEM-L), a framework which leverages real-time sensor-derived 3D geometric representations of an environment to extract information about line of sight between radios and attenuating walls/obstacles in order to accurately predict received signal strength (RSS). 
    Our data-driven approach combines the strengths of well-known models of signal propagation phenomena (e.g. shadowing, reflection, diffraction) and machine learning, and can adapt online to new environments. 
    We demonstrate the performance of PropEM-L on a six-robot team in a communication-restricted environment with subway-like, mine-like, and cave-like characteristics, constructed for the 2021 DARPA Subterranean Challenge.
    Our findings indicate that PropEM-L can improve signal strength prediction accuracy by up to 44\% over a log-distance path loss model.
\end{abstract}

\IEEEpeerreviewmaketitle

\section{Introduction}

% \ph{communication is important and challenging}
Motivated by lunar and planetary exploration \citep{titus2021roadmap,agha2019robotic},
% and disaster mitigation \citep{murphy2009mobile}, 
we consider the robotic exploration of large-scale and unknown subterranean environments. The increased coverage and redundancy offered by a team of robots can improve exploration performance, relative to a single robot. Multi-robot teams benefit from the collaboration and cooperation offered by inter-robot communication \citep{yan2013survey}. However, harsh subterranean environments typically have limited communication infrastructure, meaning robots cannot rely on wireless access points for communication. In addition, communication signals see significant degradation due to the scale of the environment, winding passages without line of sight, and obstacles.

\begin{figure}
    \centering
    \includegraphics[width=\columnwidth]{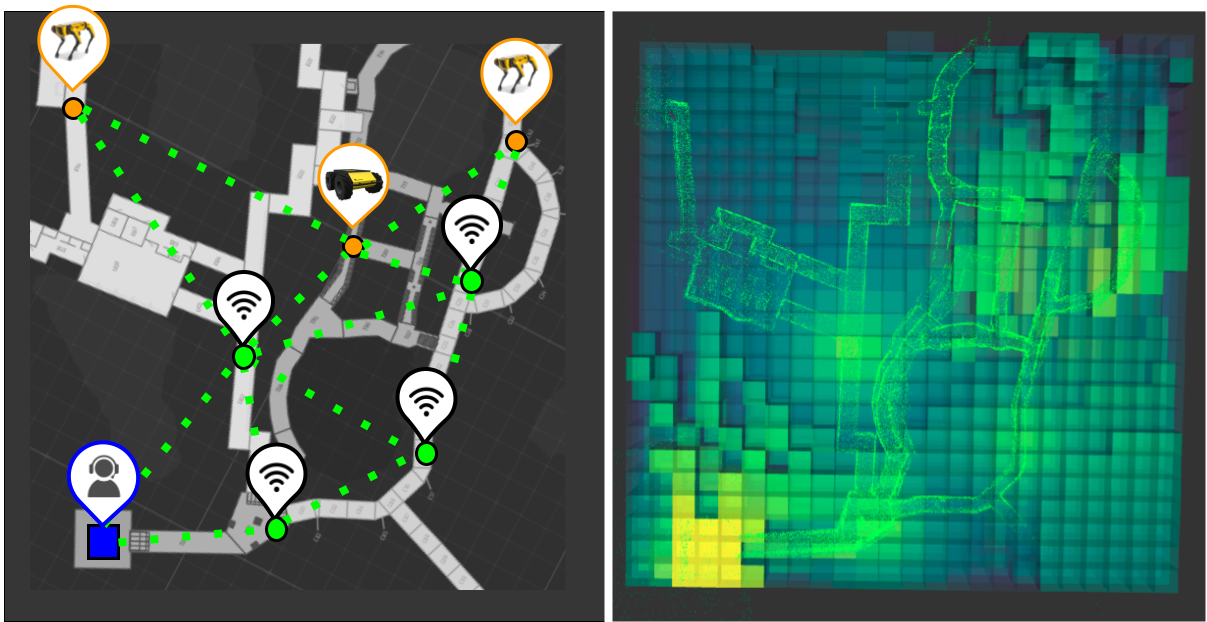}
    \caption{\textbf{Left:} Network of exploring robots, stationary relay radios, and human supervisor base station. The network provides communication during exploration, disaster mitigation, or search and rescue and is visualized over a ground truth map of the 2021 DARPA Subterranean Challenge course.
    \textbf{Right:} 3D map of connectivity predicted by PropEM-L, learned from sparse point cloud data, which can be used to enable communication-aware exploration.}
    \label{fig:ssm_grid}
\end{figure}

% \ph{RSS prediction is useful}
Modeling received signal strength (RSS) and how the environment affects it is useful for communication-aware exploration, during which robots autonomously maintain \citep{stump2008connectivity}, restore \citep{clark2021queue}, and/or improve \citep{vaquero2020traversability} connectivity. 
As shown in Fig. \ref{fig:ssm_grid}, we consider the setting of $n$ robots exploring a subterranean environment who must convey their findings to a stationary base station at the entrance of the cave/tunnel, and can relay data through stationary radios which are deployed during exploration \citep{aghamohammadi2020deployable}. 
Predicting whether robot-to-base communication is available (through one or more hops) affects decisions made by the robot, for example when and where to deploy static radios.
Additionally, understanding where connectivity is available improves situational awareness for the remote human supervisor and can assist in centralized planning and task allocation \citep{banfi2016asynchronous,otsu2020supervised}.
Because the signal strength depends on distance and the environment, accurate prediction can also aid in radio signal source-seeking \citep{fink2010online}, multi-robot localization \citep{cao2018dynamic}, and distributed task-planning \citep{vander2019mars}.
\textcolor{black}{
However, the mobility of the robots and lack of \textit{a priori} maps presents new challenges for accurate RSS prediction.
}

\subsection{Related Work}

% https://comrob.fel.cvut.cz/papers/mesas20comms.pdf

% \ph{RSS prediction}
The communication challenges associated with multi-robot exploration have drawn increased attention \citep{stump2008connectivity,robuffo2013passivity,rooker2007multi,vaquero2018approach,pei2013connectivity}. These works are largely concerned with maintaining connectivity, which requires models for estimating and/or predicting connectivity. Existing work simplifies the inter-robot communication model to a deterministic communication radius \citep{pei2013connectivity}, or predicts a probability of connectivity based solely on distance \citep{stump2008connectivity}. These simple models break down if, for instance, there is a metal wall within the communication radius which blocks the signal. For this reason, Miyagusuku \emph{et al.} seek to capture the role of the environment through a learned model of signal attenuation \citep{miyagusuku2016improving,miyagusuku2018precise}. Their data-driven approach is well-suited for repeated operation in the same environment, but for exploration of an unknown and dynamic environment an online method is advantageous.

% \ph{online RSS prediction}
The authors of \citep{fink2010online} demonstrate an online method for modeling RSS during single-robot exploration. While they present a model which captures the size and location of obstacles (for example, a 1m thick wall 4m from the transmitter), they defer the discussion of estimating these parameters.
\textcolor{black}{Quattrini Li \textit{et al.} propose a Gaussian Process (GP)-based method to build a communication map from measurements taken by multiple robots \citep{quattrini2020multi}.}
% The computational complexity of training GPs scales poorly with the number of samples, leading us to consider other methods.
\textcolor{black}{These works also focus on 2D operations, and give little attention to the challenge of modeling the propagation environment given partial maps in the form of 3D point clouds.}

% \ph{point clouds}
% \textcolor{black}{A fundamental difference between our work and the prior work mentioned thus far is that we address the challenge of propagation environment modeling given partial maps in the form of 3D point clouds. \citep{quattrini2020multi} simplify this problem by estimating the number of walls given a 2D grid map, while \citep{miyagusuku2016improving,miyagusuku2018precise} simplify the problem by assuming static transmitters in a static environment.}
Previous work in the ecology community has considered extracting relevant information from 3D point cloud data. The authors of \citep{carr2018individual} and \citep{kubelka2020radio} use point clouds to determine the location and thickness of trees in a dense forest, which could theoretically feed into the model presented in \citep{fink2010online}. Recently, the authors of \citep{egi2019machine} extend this idea to use learning-based methods on 3D point cloud data to predict the affect of tree canopies on signals propagating between static wireless communication towers. 
% \ph{terrain models}
Precise digital terrain models have also been used to inform signal propagation models. 
\citep{filiposka2011terrain} extends a network simulator with awareness of the topography, while \citep{santra2021experimental} validates terrain-aware signal strength models with experimental data. Recently, the authors of \citep{masood2019machine} demonstrate the capability of neural networks to learn from known digital terrain models to improve cellular network design.
Our work extends this method to be suitable for a dynamic network of mobile, exploring robots.

% most robot connectivity models don't use the environment
% data-driven approaches which use the environment need a priori maps
% online-learned models perform better (without the environment)
% point cloud data can be used to extract information about the environment
% terrain models add info about the environment to connectivity models

% this paper does feature extraction like carr, kubelka, egi, filiposka
% also does data-driven learning like miyagusuku
% but does it online like fink

\subsection{Contributions and Highlights}

To enable communication-aware multi-robot exploration in unknown environments, we propose Propagation Environment Modeling and Learning (PropEM-L), a framework for signal strength prediction which learns the effect of the environment on attenuation. The contributions and highlights of our work are:
\begin{enumerate}
    \item \ph{Propagation Environment Modeling} We propose PropEM, which leverages sparse 3D geometric representations of the environment (e.g. LiDAR point clouds) to extract features of the physical space which affect signal propagation, including line-of-sight visibility, shadowing due to obstacles, reflection, and diffraction. 
    \item \ph{Learning} We validate PropEM in conjunction with conventional data-driven approaches to RSS prediction which rely on linear regression. We then propose a neural network-based approach, PropEM-L, which significantly improves prediction accuracy relative to a log-distance path loss model and can estimate RSS within a few decibels.
    \item \ph{Deployment} We evaluate the performance of PropEM-L experimentally on a dynamic network of 13 autonomously-deployed static radios, one base station, and six robots exploring an underground environment. Via online learning, we demonstrate that PropEM-L can adapt to challenging new environments and construct signal strength maps.
\end{enumerate}

\begin{figure}
    \centering
    \includegraphics[width=\columnwidth]{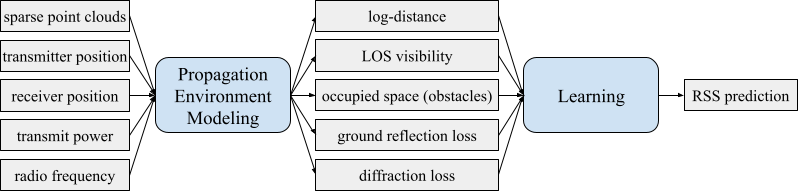}
    \caption{PropEM-L is a framework for predicting the signal strength between two radios, given their positions, which learns from the geometry of their environment. PropEM acts as an encoding layer from raw sensor data to specific features which are relevant to propagation modeling.
    }
    \label{fig:propeml_diagram}
\end{figure}

\section{Preliminaries}

% \ph{free space path loss}
Radio signal propagation is a multi-scale process where received signal strength is a function of distance between the transmitter and receiver, shadowing due to obstacles, and multi-path phenomena that result from reflections and refractions.
Free space path loss is a first-order model which quantifies the expected attenuation in an obstacle-free environment. It is typically modeled as a logarithmic function of distance $d$ given by
\begin{equation}
\label{eq:simple_pathloss}
    PL_{dB} = PL(d_0)_{dB} + \eta 10 \log_{10} (d / d_0).
\end{equation}
$PL(d_0)_{dB}$ is the reference path loss in dB at a known distance $d_0$ and $\eta$ is the path loss exponent which captures how quickly the signal falls off and typically takes on different values in different environments: in free space $\eta=2$, in indoor areas with line-of-sight $1.6 \leq \eta \leq 1.8$, in urban outdoor areas $2.7 \leq \eta \leq 3.5$, and in shadowed urban outdoor areas $3 \leq \eta \leq 5$ \citep{palacio2011radio}.

% \ph{shadowing}
Especially for long-ranges, the effect of shadowing is often captured in this model by the addition of a log-normal random variable with zero mean. Shadowing is complex to model as it depends on the environment itself, and second-order models (capturing path loss and shadowing) are usually determined by fitting samples taken within a specific environment.

% \ph{reflection}
Third-order models capture additional information about reflected and diffracted signals. The two-ray model is commonly used to capture the constructive or destructive interference caused by signals which bounce off the ground, and depends on the height of the antennas at the transmitter and receiver ($h_{tx}$ and $h_{rx}$). The two-ray model \citep{schwengler2016wireless} is given by
\begin{equation}
    PL = \frac{G_{tx} G_{rx} \lambda^2}{(4 \pi d)^2} \left(1 + \Gamma \exp \left(\frac{2 i \pi (l' - l)}{\lambda} \right) \right)^2
\end{equation}
\begin{equation}
    l' = \sqrt{\left(d^2 + (h_{tx} + h_{rx})^2 \right)}
\end{equation}
\begin{equation}
    l = \sqrt{\left(d^2 + (h_{tx} - h_{rx})^2 \right)}
\end{equation}
where $G_{tx},G_{rx}$ are the transmitter and receiver antenna gains. 
$\lambda$ is the wavelength of the radio signal. 
$\Gamma$ is the ground reflection coefficient, and we assume perfect reflection ($\Gamma$ = -1) for simplicity. 
$\frac{G_{tx} G_{rx} \lambda^2}{(4 \pi d)^2}$ also models the free space path loss, therefore the isolated \textit{reflection loss} can be modeled by
\begin{equation}
\label{eq:reflection_loss}
    RL = \left(1 - \exp \left(\frac{2 i \pi (l' - l)}{\lambda} \right) \right)^2
\end{equation}
where negative values indicate destructive interference and positive values indicate constructive interference. Note that there exist six-ray and ten-ray models which account for additional reflections, but we limit our attention to the two-ray model to avoid the additional complexity \citep{schwengler2016wireless}.

% \ph{fresnel zone}
Given knowledge of the three-dimensional space surrounding the transmitter and receiver, we can also model the Fresnel zone which is a series of ellipsoids defining the area between the transmitter and receiver where obstructions may cause interference. 
The radius of the first Fresnel zone at a given distance $d_1$ from the transmitter and $d_2$ to the receiver can be approximated by
\begin{equation}
    r = \sqrt{\frac{d_1 d_2 \lambda}{d_1 + d_2}}
\end{equation}
making the simplifying assumption that the distance between the radios is much larger than the radius \citep{schwengler2016wireless}.
Even given a clear line-of-sight path, obstacles within the first Fresnel zone can have a significant impact on signal strength \citep{rappaport1996wireless}. 

% \ph{diffraction}
For a single object, like a hill or boulder, the attenuation caused by diffraction can be estimated by treating the obstruction as a diffracting knife edge where the additional \textit{diffraction loss} is given by
\begin{equation}
\label{eq:diffraction_loss}
    DL_{dB} = 20 \log_{10}(|F(v)|).
\end{equation} 
$F(v)$ is the Fresnel integral, a function of the Fresnel-Kirchoff diffraction parameter $v$, defined as
\begin{equation}
    v = h \sqrt{\frac{2 (d_1 + d_2)}{\lambda d_1 d_2}}
\end{equation} 
where $h$ is the relative height of the obstruction (i.e. 0 if exactly in the line of sight). An approximate solution for Eq. \ref{eq:diffraction_loss} is given in \citep{filiposka2011terrain}.
As demonstrated by Filiposka and Trajanov, the combination of the two-ray model and the knife-edge model can predict values within a few dB of the ground truth and can identify isolated weak-reception areas well \citep{filiposka2011terrain}.
% other models (6-ray, 10-ray) exist but because we're raytracing in 3D voxel space each adds another layer of computation and we want to be able to predict in real-time (online), so it's advantageous to use the simplest models which work.

% Third-order models capture small-scale fading as well, and while complex ray-tracing methods can be used to model the effect of bounced signals, they are more commonly captured through the addition of a Rician random variable (for line-of-sight signals) or a Rayleigh random variable (for non-line-of-sight signals).
% \begin{equation}
%     PL = PL(d_0) + \eta 10 \log_{10} (d / d_0) + \begin{cases} \mathcal{N}(0,\mu), \\ f(x, x^s) \end{cases} + \begin{cases} \epsilon_{LOS}, \\ \epsilon_{nLOS} \end{cases}
% \end{equation}

\section{Propagation Environment Modeling}
\label{sec:feature_extraction}

% \ph{motivation}
In this section we introduce PropEM, as depicted in Fig. \ref{fig:propeml_diagram}, which serves as an encoding layer from raw point cloud data to physically-relevant context of the propagation environment between two radios. In spacious environments with multiple robots equipped with 3D LiDAR taking frequent measurements (e.g. 10Hz), the aggregated point cloud capturing the physical space around them is constantly changing in size and can quickly exceed millions of points. Learning on the raw data would be challenging; the reduction in feature space offered by PropEM is advantageous.

\ph{Occupancy grid}
PropEM creates a discretized representation of the world via OpenVDB, an open source library for the efficient storage and manipulation of sparse volumetric data in three-dimensional grids \citep{museth2013vdb}. Each voxel in this grid is associated with the probability that the space it represents is occupied. We store probabilities as log-likelihoods so that performing Bayesian updates on a voxel amounts to addition and subtraction.

\begin{figure}
    \centering
    \includegraphics[width=\columnwidth]{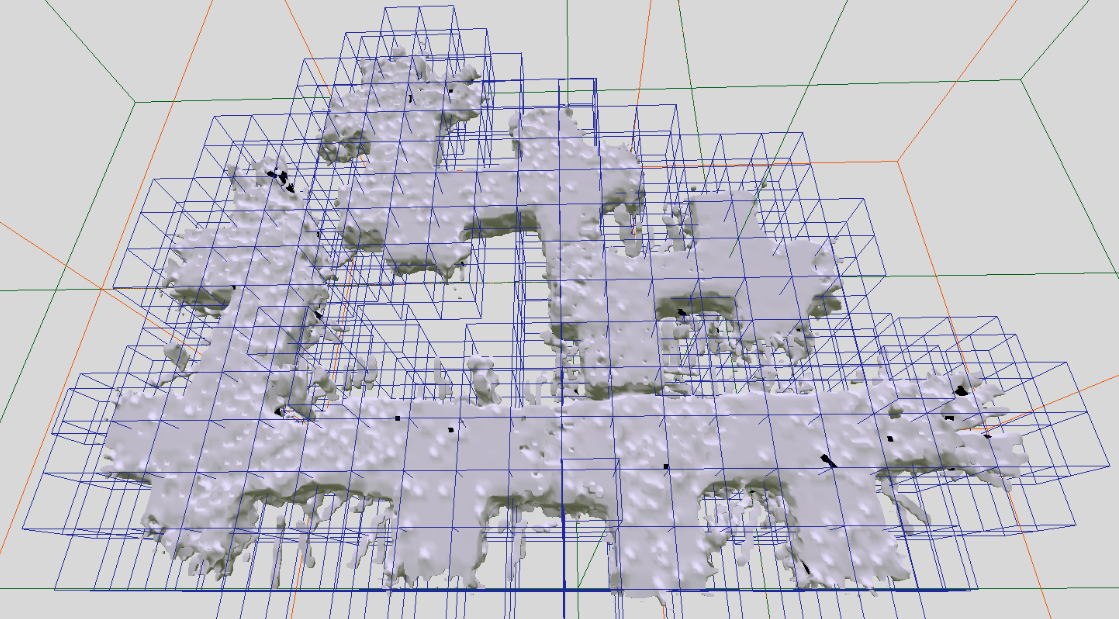}
    \caption{3D occupancy grid constructed for predicting line of sight.}
    \label{fig:occupancy_grid}
\end{figure}

\begin{figure}
    \centering
    \includegraphics[width=\columnwidth]{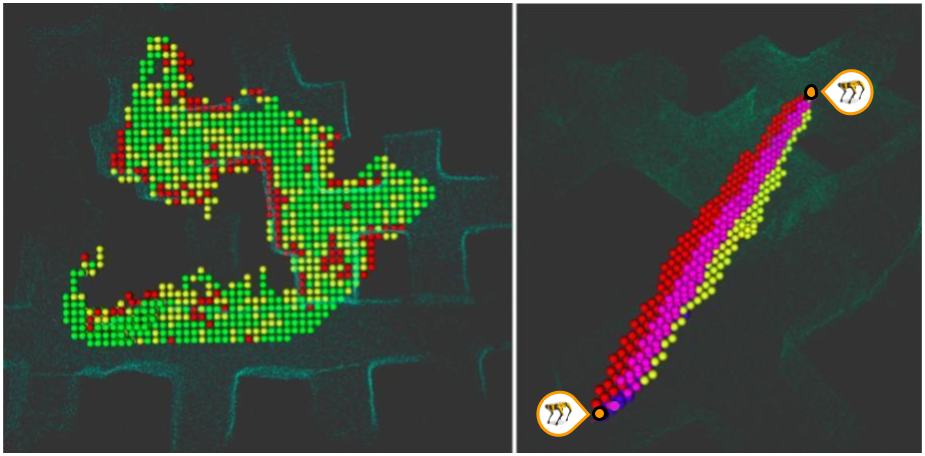}
    \caption{\textbf{Left:} Occupied (red), maybe occupied (yellow) and free (green) voxels intersected during raytracing. \textbf{Right:} Voxels which comprise the first Fresnel zone between two radios (black), where colors help illustrate perspective.}
    \label{fig:raytracing}
\end{figure}

% \ph{updating occupancy grid}
At each time $t$ that a robot takes a LiDAR scan, we associate the robot's estimated pose, $x_r(t)$, and the scan, $s_r(t)$. Localization is performed via a pose-graph SLAM algorithm called LAMP, designed for the large-scale exploration of perceptually-degraded environments \citep{ebadi2020lamp,palieri2020locus}. 
PropEM performs raytracing from the robot's estimated position to each point in the point cloud representation of the scan. Voxels along these rays have an increased likelihood of being unoccupied (as the light ray did not intersect an obstacle) while voxels at the points have an increased likelihood of being occupied. 
Following the approach presented in \citep{raytracingtutorial}, PropEM performs this Bayesian update for all robots and all scans in a stationary, shared frame. The result is a 3D occupancy grid as shown in Fig \ref{fig:occupancy_grid}. Likelihoods continue to update as robots re-visit an area, such that we can account for dynamic obstacles.

\ph{Line of sight}
Given an arbitrary pair of positions representing a transmitter and receiver, we can use this occupancy grid to determine the propagation environment which would affect the signal passing between them. To trace the LOS ray, PropEM uses digital differential analyzers (DDAs), a technique from computer graphics used to interpolate between two points. 
We traverse the voxels between the transmitter and receiver, and classify them as free, occupied, maybe occupied, or unknown. This classification is done by thresholding the log-likelihoods associated with each voxel (see Fig. \ref{fig:raytracing}). The number of total and occupied voxels informs our calculation of free space path loss and shadowing.

\ph{Reflection and diffraction}
To consider third-order effects like reflection and diffraction, we follow the approach presented in \citep{filiposka2011terrain} to consider a model of the terrain (in our case the occupancy grid representing the partially known environment) as it causes reflections from the ground and as it intersects with the first Fresnel zone. We calculate the reflection loss $RL$ given by Eq. \ref{eq:reflection_loss} as a function of the transmitter and receiver locations. Then, incrementally along the line-of-sight path, we traverse voxels horizontally and vertically in the robot's frame of reference to check the first Fresnel zone clearance (see Fig. \ref{fig:raytracing}). Along each direction we calculate the Fresnel-Kirchoff diffraction parameter $v$ determined from the occupancy grid, and use the worst case $v$ to approximate the diffraction loss $DL$ given by Eq. \ref{eq:diffraction_loss}.

% \ph{model outputs}
By maintaining and updating the occupancy grid representation internally, PropEM can determine for an arbitrary pair of radio positions the distance, visibility, and obstacles between them as well as the ground reflection loss and diffraction loss from the cave/tunnel walls. In the next section, we discuss how these features are used in RSS prediction.

\section{Learning}
\textcolor{black}{PropEM outputs the physically-relevant features of the propagation environment, and we use these features to perform data-driven radio signal strength prediction.}

\subsection{Field Test Data}
% \ph{dataset}
We performed extensive field testing in an underground limestone mine in Kentucky and captured about 80 minutes of autonomous exploration with 1 base station radio, 6 ground robots (3 legged Boston Dynamics Spot robots and 3 wheeled Clearpath Robotics Husky robots), and 13 communication relay radios which were autonomously deployed by the ground robots when signal-to-noise ratio (SNR) fell below a desired threshold. 
All radios were MIMO StreamCaster 4000-series radios from Silvus Technologies \citep{ginting2021chord, saboia2022achord}.
Exploration spanned about 300 meters in each of the $x$ and $y$ directions, and we collected 1.3 million datapoints capturing transmitter position, receiver position, frequency, noise, transmitter power, RSS, SNR, loss rate, etc \citep{agha2021nebula}.

% \ph{Data augementation and sanitization}
Although we can only collect signal strength data between pairs of radios which have connectivity, we have location estimate data for all radios. 
For pairs of radios which have certainly lost connectivity, meaning neither of the robots report RSS measurements at a given location for a two-minute period, we augment the dataset following the approach presented in \citep{miyagusuku2016improving}, where the received signal strength is inferred to be the noise floor of the radios (here -94dB).
% and the SNR is inferred to be 0. 
% We observe, however, instances when the signal strength data is briefly unavailable for other reasons (this could be due to the radio API query failing to respond in a timely enough matter as the network size scales, or the data being dropped between its collection at the remote radio pair and its processing at the base). Thus we limit the inclusion of these augmented data points to pairs of radios which have no measurements during a two-minute period and no other measurements at the reported distance.
To ensure synchronization between the location estimate and the signal strength measurement, we remove any data for which the two were observed more than one second apart.
Additionally, we observe a limited number of outliers caused by localization failure which are easily removed for static radios whose location estimates should never vary more than a few meters.

% I also had to remove the last 10 minutes of the 96 minute rosbag because some of the dropped radios were moved before they were turned off.
% For competition data, have to remove txpwactualdBm == -1111 which indicates bad response from API.
% there also must be links in both directions (so excluding scom18->scom20 for comp2)

% \ph{validating PropEM}
As a preliminary step in validating PropEM, we note the correlation between the features PropEM outputs and the RSS measurements. Table \ref{tab:feature_correlation} captures the Pearson correlation coefficient of data collected during field testing, which illustrates that signal attenuation is correlated with distance and the number of \textit{not-free} voxels (occupied, maybe occupied, or unknown), followed by the additional loss predicted by the two-ray ground reflection model.

\begin{table}
    \centering
    \caption{Correlation between PropEM output features and RSS.}
    \begin{tabular}{c|c}
        Feature & Correlation with RSS (dB) \\
        \hline
        Log distance (m) & -0.84 \\
        Line of sight & 0.39 \\
        Not-free voxels (\#) & -0.78 \\
        Reflection loss (dB) & 0.58 \\
        Diffraction loss (dB) & 0.34 \\
    \end{tabular}
    \label{tab:feature_correlation}
\end{table}

\begin{figure}
    \centering
    \includegraphics[width=\columnwidth]{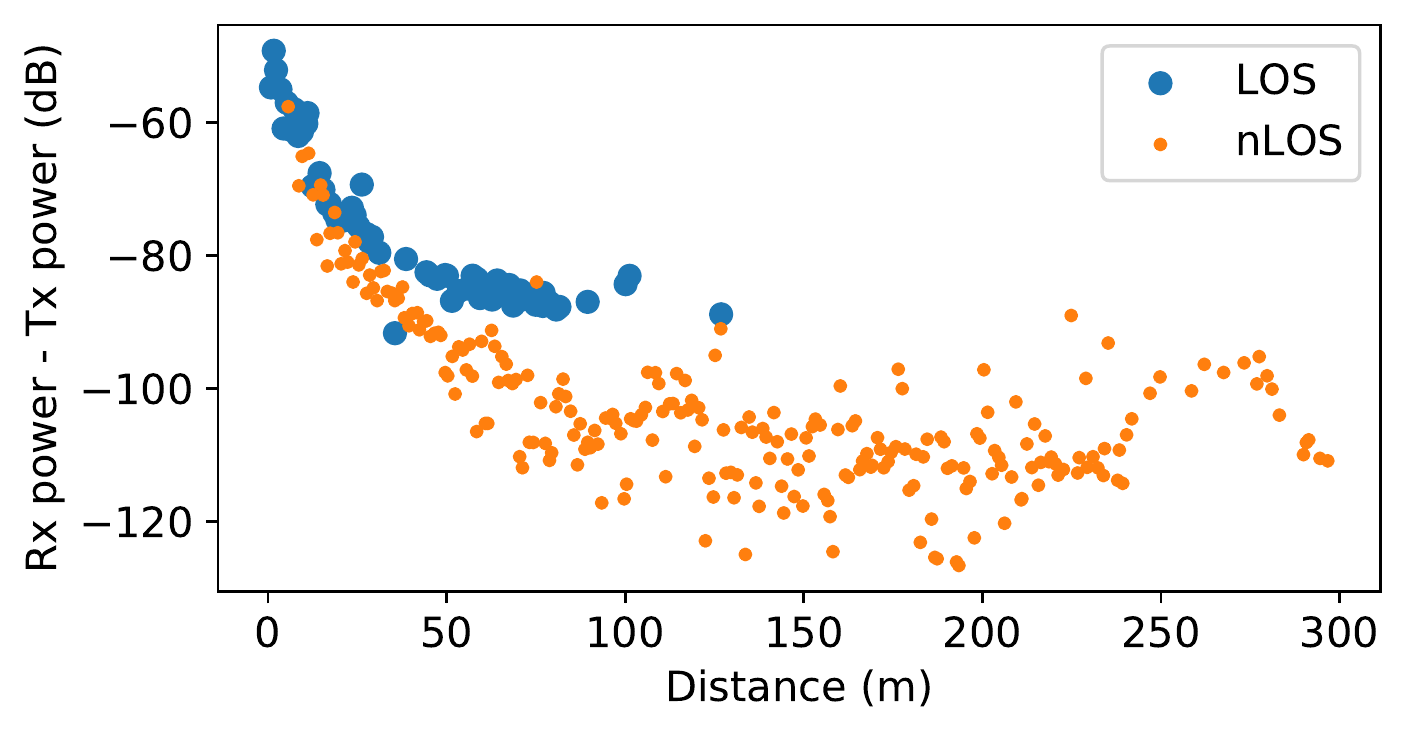}
    \caption{Signal attenuation plotted against distance. Measurements at each distance are averaged. Line-of-sight (LOS) links see less degradation than non-line-of-sight (nLOS) links.}
    \label{fig:los_nlos}
\end{figure}

\subsection{Conventional Propagation Models}
\begin{figure}
    \centering
    \includegraphics[width=\columnwidth,trim={0 40 0 40}]{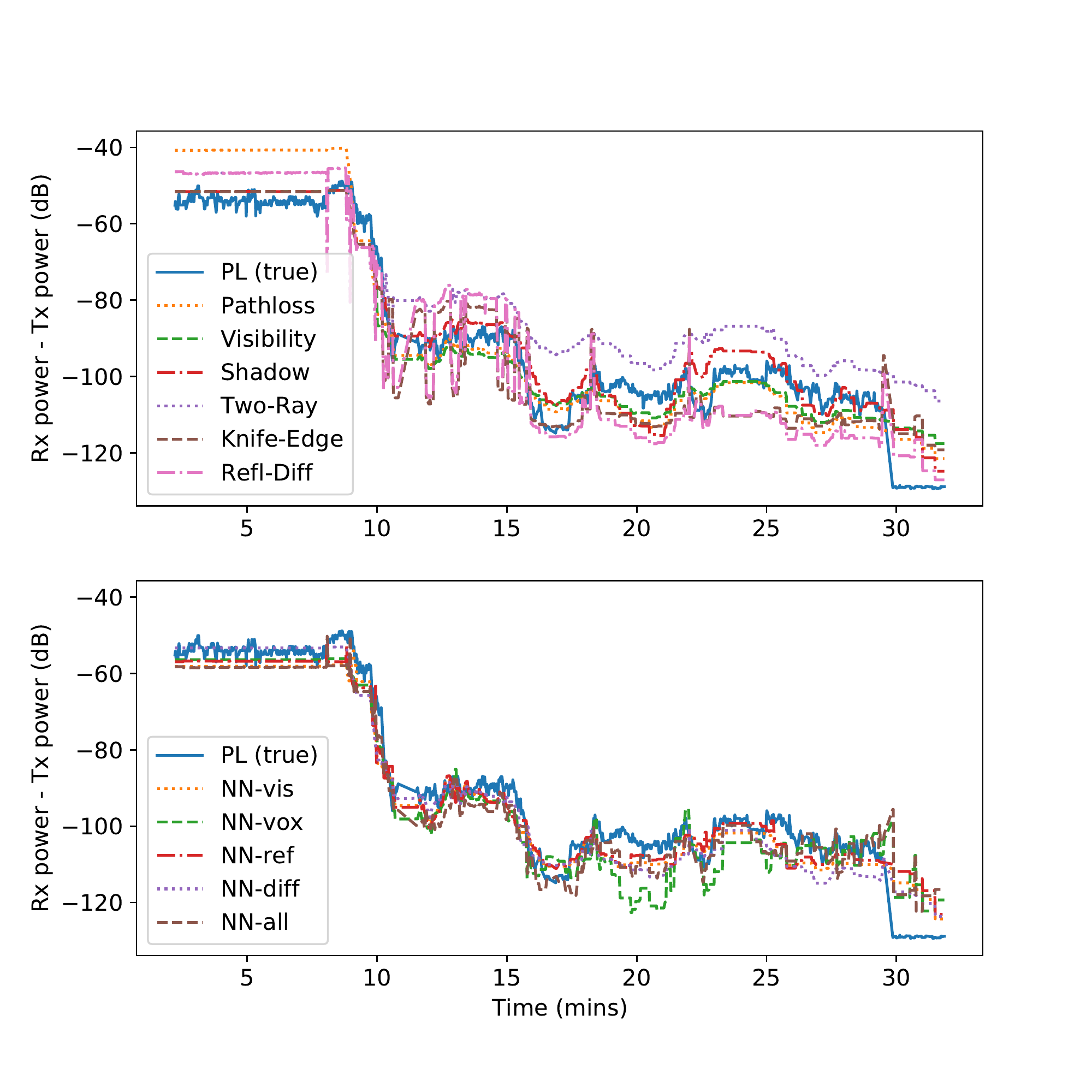}
    \caption{Example performance of the learning models visualized against ground-truth attenuation ($PL$). \textbf{Top:} conventional data-driven propagation models. \textbf{Bottom:} neural-network based prediction. These plots depict the link between a single exploring robot and the base station.}
    \label{fig:model_comparison}
\end{figure}

To demonstrate PropEM as part of an RSS prediction framework, and as baselines for comparison, we implement the standard log-distance path loss prediction model and several extensions of this model. Table \ref{tab:results} presents a summary of the accuracy of each model, and Fig. \ref{fig:model_comparison} offers a snapshot of their performance.

\ph{Simple Path Loss}
Using linear regression, we fit the parameters of the simple path loss model given by Eq. \ref{eq:simple_pathloss}. We abbreviate $PL(d_0)_{dB}$ at $d_0=1$m to $PL_{d0}$ in this and the following sections. This regression gives $PL_{d0} = 14.84$ and $\eta = 4.73$. 

\ph{Visibility}
Motivated by the distinctly different curves observed in the data, as demonstrated in Fig. \ref{fig:los_nlos}, we implement a model which incorporates visibility. We use linear regression to fit two distinct lines to Eq. \ref{eq:simple_pathloss} depending on whether there is line of sight, which gives $PL_{d0}^{\textrm{LOS}} = 36.5$, $\eta^{\textrm{LOS}} = 2.75$, $PL_{d0}^{\textrm{nLOS}} = 13.72$ and $\eta^{\textrm{nLOS}} = 4.81$.

\ph{Shadowing Heuristic}
Considering not just visibility but the amount of occupied space, we use linear regression to approximate the dB of attenuation per meter of not-free voxels.
Similar to the approach presented in \citep{bruggemann2009adaptive}, this model adds an additional attenuation of 0.16 dB/m to the output of the path loss model with LOS parameters.

\ph{Two-Ray}
This ground reflection model subtracts $RL_{dB}$, given by Eq. \ref{eq:reflection_loss}, from the simple path loss model using LOS parameters.

\ph{Knife-Edge}
This diffraction model subtracts $DL_{dB}$, given by Eq. \ref{eq:diffraction_loss}, from the simple path loss model using LOS parameters.

\ph{Reflection-Diffraction}
As presented in \citep{filiposka2011terrain}, this model combines the effects of two-ray ground reflection and knife-edge diffraction. For a maximum $v \leq -0.8$, the Fresnel zone is considered clear and this model reduces to the two-ray model. Otherwise, the diffraction loss is subtracted from the two-ray model with the modification that any constructive reflective interference is ignored.

% Where the Filiposka model 
% if v <= -0.8, GdB >= 0 (no diffraction loss): pathloss + two-ray
% else: worse(pathloss, pathloss+two-ray) + diffraction loss

%  scom-spot3, scom-base1 (sep18d)

% \ph{performance}
Of these models, the visibility model achieves the most accurate prediction, as shown in  Fig. \ref{fig:model_comparison} (top). 
The two-ray, knife-edge, and reflection-diffraction models are very sensitive to small differences in antenna height, distance between transmitter and receiver, and precise location of neighboring walls. Because PropEM relies on sparse point clouds, uncertain location measurements, and discretized space (voxels are up to 1.5m wide), we see rapid changes in the predicted diffraction loss as new voxels are explored or the robot's position changes even slightly. This sensitivity is likely the reason the more parsimonious models actually perform better.

\subsection{Learned Propagation Models (Neural Networks)}
% \ph{motivation for NNs}
As implemented, the conventional models make a few simplifying assumptions: the simple path loss model assumes no obstacles, the visibility and shadowing heuristic models assume all obstacles have the same material properties, the two-ray model assumes a flat reflective terrain, and the knife-edge model assumes the worst-case from obstructions. While these models have the advantage of being interpretable and grounded in the physics of how radio signals propagate, non-linear combinations of these features, learned via stochastic gradient descent, can capture subtleties in the environmental dependence which are inadequately captured in the simpler models.

% \ph{NN structure}
We introduce five fully-connected neural networks (NN), each with the same simple structure: an input layer, a sixteen-unit hidden layer with restricted linear activation function, and an output layer which predicts attenuation $PL_{dB}$. Each NN learns to minimize the L2 loss between the predicted attenuation and the measured attenuation, using the Adam optimizer to train for 100 epochs or until convergence. The field test data from the limestone mine is normalized such that all features have zero-mean and unit standard deviation. 30\% of the data is withheld from training, and the remainder is divided randomly into batches of size 2048.

% \ph{NN models}
Each NN takes a subset of the features which PropEM outputs, as described below:
\begin{itemize}
    \item \textbf{NN-vis} uses logarithmic distance and LOS, which is broken into two boolean input features: \textit{strictly visible} indicates all the voxels between the transmitter and receiver are free; \textit{strictly not visible} indicates there is at least one occupied voxel.
    \item \textbf{NN-vox} uses logarithmic distance and the number of occupied, maybe occupied, free, and unknown voxels. 
    \item \textbf{NN-ref} uses logarithmic distance and reflection loss.
    \item \textbf{NN-diff} uses logarithmic distance and diffraction loss.
    \item \textbf{NN-all} combines the features of NN-vox, -vis, and -diff.
\end{itemize}
% NN-vis converged in 40 epochs.
% NN-vox converged in 66 epochs.
% , and did not converge.
%  , and converged in 5 epochs.
% , and converged in 90 epochs.

\begin{table}
    \centering
    \caption{Mean absolute error (dB) for each model (offline)}
    % during field test data from the limestone mine and each run of the DARPA SubT Challenge.}    
    \begin{tabular}{c|c|c|c|c}
        Learning Model & Field Test & Day 1 & Day 2 & Day 3 \\
        \hline \hline
        Simple path loss & 6.73 & 15.87 & 15.37 & 13.0 \\
        Visibility & \textbf{6.5} & 13.93 & \textbf{14.14} & \textbf{12.57} \\
        Shadowing Heuristic \citep{bruggemann2009adaptive} & 6.61 & 14.12 & 14.55 & 12.88 \\
        % \hline 
        Two-Ray & 12.07 & 17.59 & 17.86 & 16.57 \\
        Knife-edge & 10.18 & \textbf{13.54} & 14.69 & 13.36 \\
        Reflection-Diffraction \citep{filiposka2011terrain} & 10.15 & 13.93 & 15.45 & 14.11 \\
        \hline 
        % NN Model 1 & 6.11 & 12.63 & 13.42 & 12.03 \\
        % NN Model 2 & 3.87 & 11.87 & \textbf{12.70} & \textbf{11.21} \\
        % NN Model 3 & 6.14 & 13.00 & 13.56 & 12.58 \\
        % NN Model 4 & 6.36 & 12.64 & 14.99 & 12.43 \\
        % NN Model 5 & \textbf{3.73} & \textbf{11.42} & 12.95 & 11.70 \\
        NN-vis & 6.11 & 12.63 & 13.42 & 12.03 \\
        NN-vox & 3.87 & 11.87 & \textbf{12.70} & \textbf{11.21} \\
        NN-ref & 6.14 & 13.00 & 13.56 & 12.58 \\
        NN-diff & 6.36 & 12.64 & 14.99 & 12.43 \\
        NN-all & \textbf{3.73} & \textbf{11.42} & 12.95 & 11.70 \\
    \end{tabular}
    \label{tab:results}
\end{table}

\begin{figure}
    \centering
    \includegraphics[width=0.9\columnwidth]{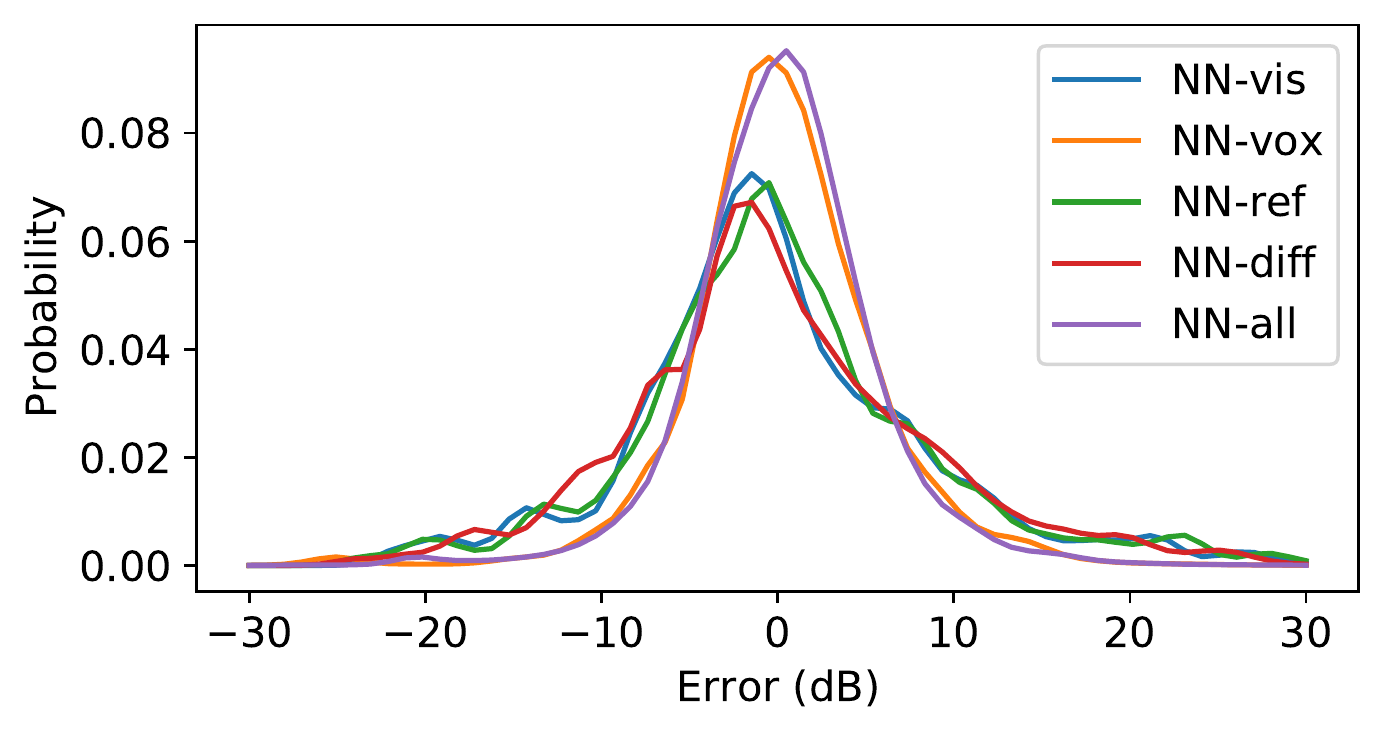}
    \caption{Error probability density functions for each NN model.}
    \label{fig:pdfs}
\end{figure}

% \ph{performance}
Table \ref{tab:results} compares the performance of each NN model, and Fig. \ref{fig:pdfs} depicts the probability density function of errors observed using each model.
NN-all saw the best performance of any method with an average error of only 3.72 dB, and is less likely to produce errors above 10dB.
To contextualize this performance, the standard deviation of $PL_{dB}$ for a pair of static radios (at a constant distance) observed in the limestone mine was up to 3.43 dB, indicating that errors in this range may be attributed to inevitable channel noise.
% ky scom6 -> scom12 std(measured_path_loss_dB) which are 57 meters apart

% \subsection{Performance in a new environment}
\section{Adapting to New Environments}

% \ph{competition environment}
One of the most important aspects of a prediction framework is the ability to generalize and transfer across different, complex environments. We collected data from two 30-minute (days 1 and 2) and one 60-minute (day 3) analog exploration missions in the course constructed for the DARPA Subterranean Challenge \citep{agha2021nebula}. This environment, shown in Fig. \ref{fig:ssm_grid} was characterized by narrow, winding passageways and had three distinct subsections: an urban environment similar to a subway station, a mine-like environment, and a subterranean cave-like environment. We had no \textit{a priori} knowledge of the course or the materials from which it was constructed.

\begin{figure}
    \centering
    \includegraphics[width=0.8\columnwidth]{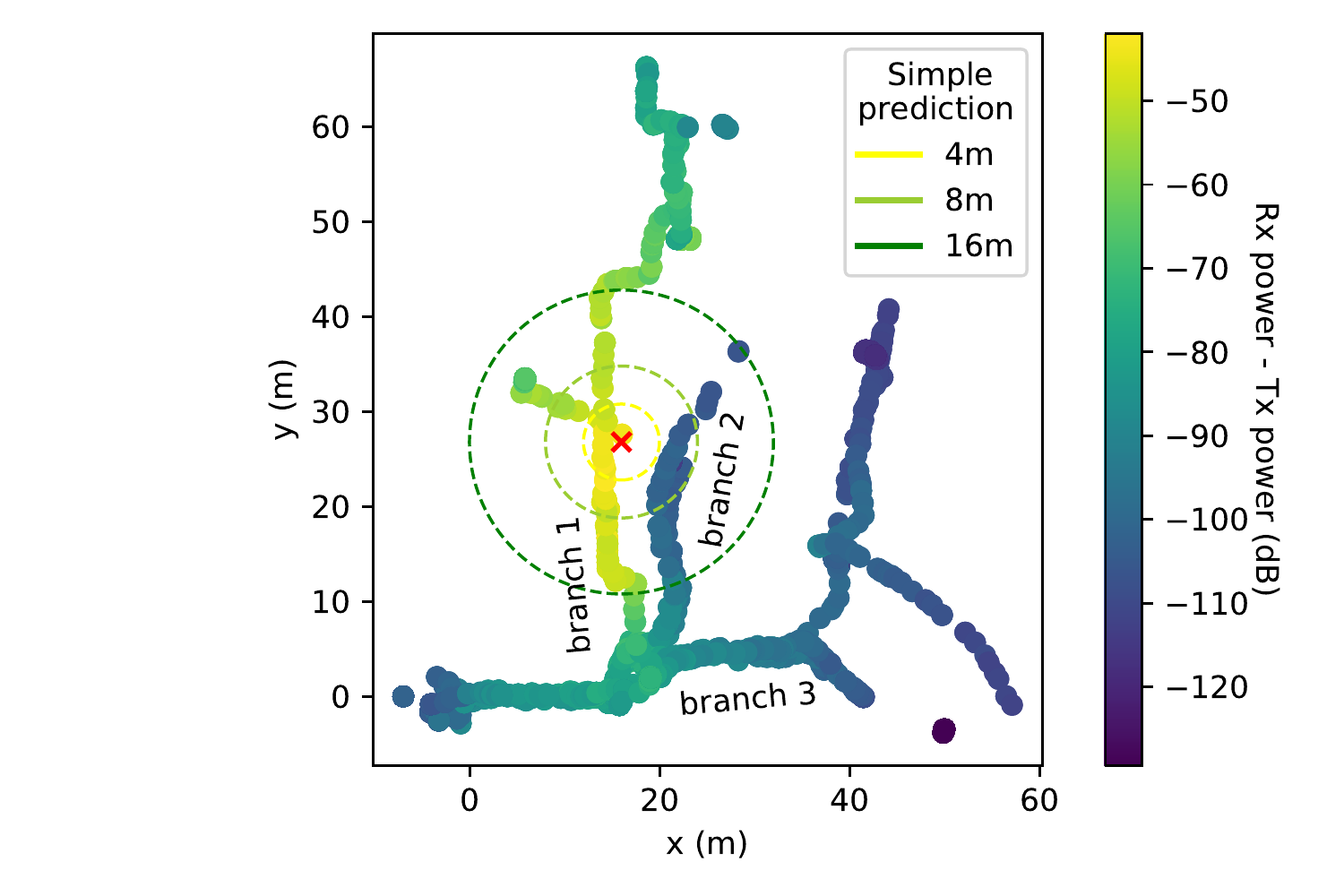}
    \caption{Spatial plot of signal strength from a static transmitter (red) in the environment depicted in Fig. \ref{fig:ssm_grid}. \textcolor{black}{A simple path loss model predicts the signal will attenuate with distance (circles around transmitter). However, branch 2 sees more degradation than predicted due to barriers between branches.}}
    \label{fig:comp3}
\end{figure}

% \ph{performance}
Table \ref{tab:results} summarizes the results of each model for each day of the competition. The average error for each model was more than 10dB, which could be due to a number of things. The scale of this environment was drastically different from that of the field test, with tunnels of ~2m width rather than 20m width. 
Additionally, the environment was constructed of different materials (e.g. thin metal panel dividers in place of thick limestone columns). These materials attenuated signal more significantly than expected, as illustrated in Fig. \ref{fig:comp3}.
As a result, a simple path loss model was unable to accurately predict signal strength, with more than 15dB of error on average during each 30-minute experiment.

\subsection{Online Learning}
To adapt to drastically different environments, PropEM-L can perform online learning to leverage the partial 3D geometric representations available during exploration. Note that for simplicity we assume in this section RSS measurements from all radios are available at the base station immediately, ignoring any networking latencies.

\ph{Linear regression}
Using the visibility model and re-fitting the parameters $\eta^{\textrm{LOS}}, PL_{d0}^{\textrm{LOS}}, \eta^{\textrm{nLOS}}, PL_{d0}^{\textrm{nLOS}}$ every $k$ minutes, we can achieve a mean absolute error of
10.99dB (21\% improvement over the offline-learned model) on day 1, 12.47dB (12\% improvement) on day 2, and 11.89dB (5\% improvement) on day 3 for $k=1$. 
\textcolor{black}{These improvements are due to the fact that the model updates the attenuation it associates with a given number of occupied voxels between a transmitter and receiver. This reflects a change in the model’s belief about the RF properties under identical geometries.
}

% \ph{inference about the environment}
Perhaps more notably, the final path loss parameters learned via linear regression can shed insight on the environment itself. The learned $\eta^{\textrm{LOS}}$ in the new environment was between 1.59-1.94, which indicates indoor areas and is significantly lower than the initial estimate of 2.75 which is typical for outdoor urban environments. Without any other information about the map, we can conclude the competition environment was more confined. Similarly, re-fitting the heuristic model increased our approximation of attenuation over not-free space from 0.16 to 0.43-0.66 dB/m. This lets us infer that the competition environment had material which blocked radio waves on average up to 4 times more than expected. This interpretability is an advantage of the conventional linear regression models over the NN models.

\begin{figure}
    \centering
    \includegraphics[width=\columnwidth]{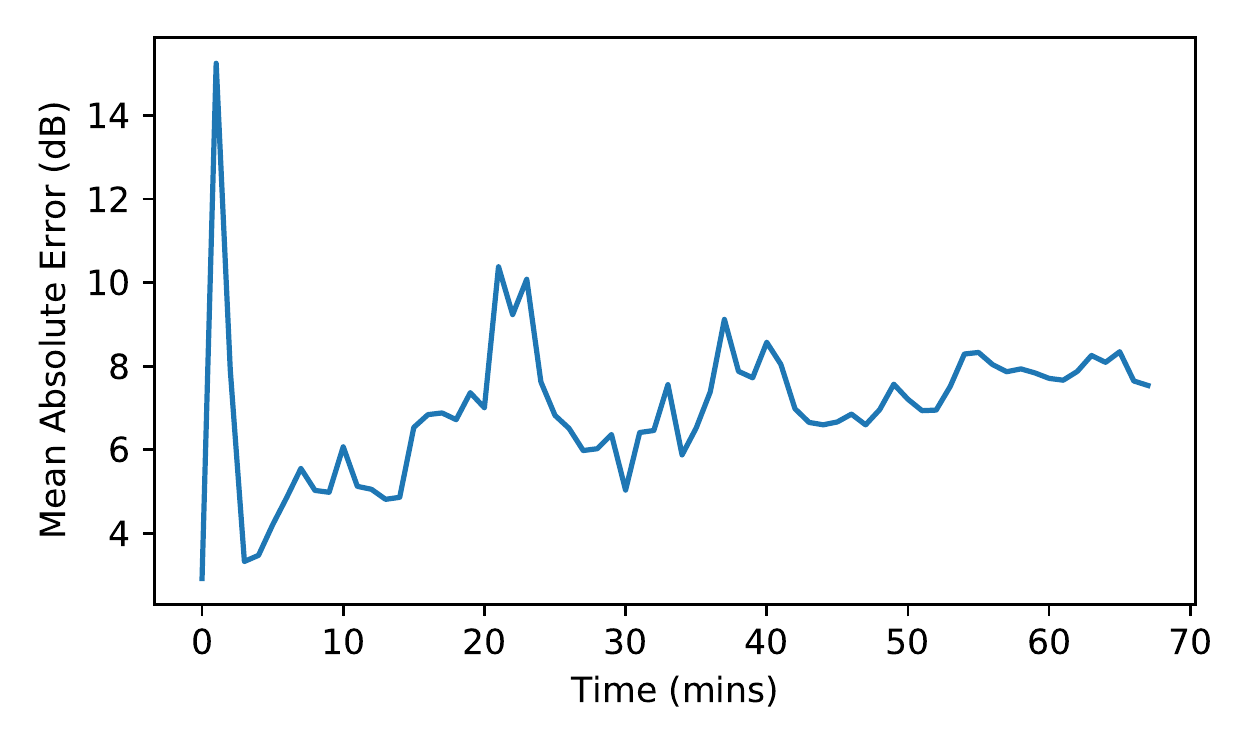}
    \caption{Performance of NN-vox trained online in the competition environment on day 3. By the final minute of exploration, mean absolute error is 7.54 dB.}
    \label{fig:online}
\end{figure}

\ph{Neural Networks}
To train the neural network component of PropEM-L online during exploration, every 60 seconds we train for 10 epochs (selected empirically) on data collected in the last minute. 
\textcolor{black}{
The number of samples used to retrain is on average 2644 (with a minimum of 498 and a maximum of 5414).
}
Starting with the model trained offline prevents overfitting to initial measurements, while over time the new environment is learned.
Fig. \ref{fig:online} shows how the mean absolute error decreases and converges over time. Over the entire duration of the competition run (day 3), NN-vox trained online can achieve a mean absolute error of 7.28 dB, improving performance over the simple path loss model trained offline almost two-fold.

\begin{table}
    \centering
    \caption{Mean absolute error (dB) for each NN model trained online.}
    \begin{tabular}{c|c|c|c}
        Model & Day 1 & Day 2 & Day 3 \\
        \hline \hline
        % NN Model 1 & 11.20 & 11.77 & 10.48 \\
        % NN Model 2 & \textbf{8.65} & \textbf{9.60} & \textbf{7.28} \\
        % NN Model 3 & 10.55 & 11.81 & 10.70 \\
        % NN Model 4 & 9.57 & 11.69 & 9.51 \\
        % NN Model 5 & 8.76 & 9.70 & 7.61 \\
        NN-vis & 11.20 & 11.77 & 10.48 \\
        NN-vox & \textbf{8.65} & \textbf{9.60} & \textbf{7.28} \\
        NN-ref & 10.55 & 11.81 & 10.70 \\
        NN-diff & 9.57 & 11.69 & 9.51 \\
        NN-all 5 & 8.76 & 9.70 & 7.61 \\        
    \end{tabular}
    \label{tab:onlineresults}
\end{table}

% \begin{figure}
%     \centering
%     \includegraphics[width=\columnwidth]{images/clean2_online_1_1b.png}
%     \caption{NN Model 2 trained online, during exploration. The model's error decreases from 54.5 dB in the first minute to 8.6 dB in the last minute.}
%     \label{fig:online}
% \end{figure}

% \begin{figure}
%     \centering
%     \includegraphics[width=\columnwidth,trim={0 0 0 32},clip]{images/online.jpg}
%     \caption{Mean absolute error (dB) of the NN-vox model trained online in the competition environment on day 3.}
%     \label{fig:online}
% \end{figure}

\subsection{Signal Strength Mapping}
PropEM-L can predict the strength of a signal received by a robot at any arbitrary location from one or multiple nearby radios. Given the location of static radios and their transmit powers, we can use our framework to construct a signal strength map, as shown in Fig. \ref{fig:ssm_grid}. PropEM-L can update this map as areas are explored or the network topology changes. Connectivity maps have a number of uses in communication-aware exploration: robots can autonomously identify areas to deploy additional relay radios \citep{saboia2022achord}, prioritize exploring new areas which are expected to offer connectivity, or navigate to areas with strong signal to prioritize data transfer and improve situational awareness for the human supervisor \citep{clark2021queue}.

\section{Conclusion}
In this work, we develop Propagation Environment Modeling and Learning (PropEM-L), a framework for signal strength prediction which learns the effect of complex, communication-restricted  environments on attenuation.
Our approach leverages sparse sensor-derived 3D geometric representations to model features of the propagation environment between a transmitter and receiver, including line-of-sight visibility, shadowing due to obstacles, reflection, and diffraction. We compare our neural network-based online learning with conventional approaches to RSS prediction, and demonstrate PropEM-L on a dynamic network of exploring robots and stationary radios in multiple communication-restricted, underground environments. Our findings indicate that PropEM-L can significantly improve RSS prediction and adapt to new environments, which will be an important aspect of transferring communication-aware exploration strategies from analog missions to real exploration missions.

Through online learning we can also infer certain things about the environment itself, which can give human supervisors (e.g. scientists and first responders) a better understanding of its scale and material properties. Formalizing this study is an interesting direction for future work.
\textcolor{black}{
Additionally, future work could consider other learning methods (e.g. Gaussian Processes), other representations (e.g. time-series RSS measurements), or other methods of encoding raw sensor data (e.g. autoencoders).
}

% Future work: moving towards distributed prediction (e.g. federated learning)

\section*{Acknowledgements}
% Acknowledgements omitted for anonymous review.
We gratefully acknowledge all members of Team CoSTAR. We also thank Belal Wang and Silvus Technologies for their hardware and technical expertise, and Kentucky Underground Storage for access to their facilities. This research was carried out at the Jet Propulsion Laboratory, California Institute of Technology, under a contract with the National Aeronautics and Space Administration (80NM0018D0004). This work was supported in part by NASA Space Technology Research Fellowship Grant No. 80NSSC19K1189.

\bibliographystyle{plainnat}
\bibliography{bib}

\end{document}